\documentclass[letterpaper]{article} 
\usepackage[draft]{aaai25}  
\usepackage{times}  
\usepackage{helvet}  
\usepackage{courier}  
\usepackage[hyphens]{url}  
\usepackage{graphicx} 
\usepackage{natbib}  
\usepackage{caption} 
\usepackage{algorithm}
\usepackage{algorithmic}

\usepackage{newfloat}
\usepackage{listings}
\DeclareCaptionStyle{ruled}{labelfont=normalfont,labelsep=colon,strut=off} 
\floatstyle{ruled}
\newfloat{listing}{tb}{lst}{}
\floatname{listing}{Listing}
\usepackage{color}

\usepackage{listings}
\usepackage{xcolor}
\usepackage{mdframed}
\definecolor{lightgraytrans}{gray}{0.8}
\colorlet{customgray}{lightgraytrans!20}
\newmdenv[
    backgroundcolor=customgray,
    linecolor=customgray,
    innerleftmargin=10pt,
    innerrightmargin=10pt,
    innertopmargin=10pt,
    innerbottommargin=10pt
]{grayquote}

\usepackage{algorithm}
\usepackage{algorithmic}
\usepackage{amsmath}
\usepackage{booktabs}
\usepackage{multirow}
\usepackage{hyperref} 
\hypersetup{colorlinks=true, linkcolor=blue, urlcolor=blue,citecolor=black}
\nocopyright

\title{Uncovering Knowledge Gaps in Radiology Report Generation Models \\ through Knowledge Graphs}

\author{
    Xiaoman Zhang, 
    Juli\'an N. Acosta,
    Hong-Yu Zhou,
    Pranav Rajpurkar
}
\affiliations{
    Department of Biomedical Informatics, Harvard Medical School, Boston, MA, USA. \\ \vspace{3pt}
    \url{https://github.com/rajpurkarlab/ReXKG}
}

\usepackage{bibentry}

\begin{document}

\maketitle

\begin{abstract}

Recent advancements in artificial intelligence have significantly improved the automatic generation of radiology reports. 
However, existing evaluation methods fail to reveal the models' understanding of radiological images and their capacity to achieve human-level granularity in descriptions. 
To bridge this gap, we introduce a system, named ReXKG, which extracts structured information from processed reports to construct a comprehensive radiology knowledge graph.
We then propose three metrics to evaluate the similarity of nodes (ReXKG-NSC), distribution of edges (ReXKG-AMS), and coverage of subgraphs (ReXKG-SCS) across various knowledge graphs. 
We conduct an in-depth comparative analysis of AI-generated and human-written radiology reports, assessing the performance of both specialist and generalist models. 
Our study provides a deeper understanding of the capabilities and limitations of current AI models in radiology report generation, offering valuable insights for improving model performance and clinical applicability.
\end{abstract}

%

\section{Introduction}

Artificial Intelligence (AI) models have recently achieved remarkable success in interpreting medical images~\cite{rajpurkar2023current,rajpurkar2022ai}. 
Among them, radiology report generation stands out as a crucial task in medical imaging, providing essential information for further diagnosis and treatment planning~\cite{liu2023systematic,reale2024vision}.
Its significance has led to a surge in research focused on developing AI models capable of generating these reports~\cite{zhang2020radiology,liu2024bootstrapping}.
However, in-depth understanding radiology report generation models' performance is a challenging yet important task for real clinical usage.

Various automated evaluation metrics have been proposed specifically for report generation, such as RadCliQ~\cite{yu2023evaluating}, FineRadScore~\cite{huang2024fineradscore}, RaTEScore~\cite{zhao2024ratescore} and GREEN~\cite{ostmeier2024green}, {\em etc}. 
These metrics have gradually approached the quality of radiologists' evaluations.
Yet, most existing metrics rely on report-to-report comparisons, which fail to fully capture a model's holistic understanding of radiological images or its capacity to match the descriptive granularity used by humans. 
For example, when a doctor mentions ``edema'' in a report, they may use nuanced modifiers such as ``moderate'', ``mild'', ``unchanged'', ``decreased'', or ``stable'' to convey precise details. In contrast, a model might not capture this level of detail or variation in terminology.
It is essential to develop evaluation methods considering the comprehensiveness of medical terminology understanding.
These insights can guide the improvement of report generation models, ensuring they are better aligned with the professional descriptions used by radiologists.

\begin{figure*}[t]
    \centering
    \includegraphics[width=1\linewidth]{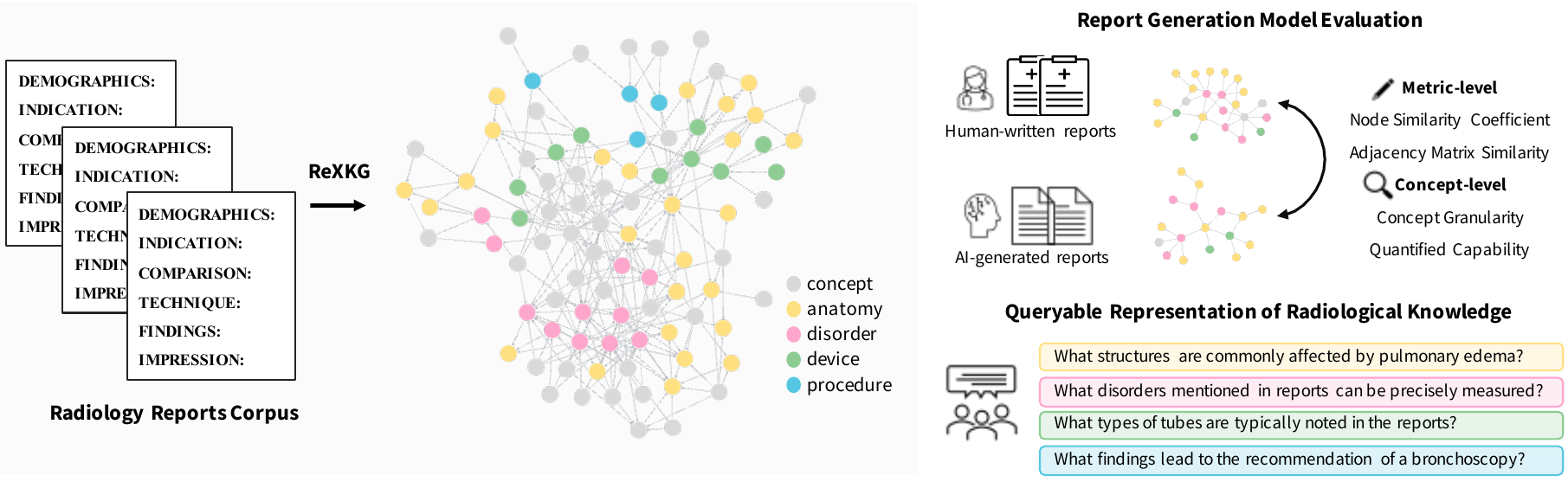}
    \vspace{-10pt}
    \caption{An illustration of \textbf{Learning from Knowledge Graph}.}
    \label{fig:teaser}
\end{figure*}

In this paper, we target assessing AI models from a different perspective by focusing on the radiological knowledge learned by the model.
To accomplish this, we introduce a system named \textbf{ReXKG}, designed to extract structured information from processed reports and construct a comprehensive radiology knowledge graph.
As shown in Figure~\ref{fig:teaser}, this graph will capture relationships between anatomical structures, pathologies, imaging findings, medical devices, and procedures, creating a rich, queryable representation of radiological knowledge.
We propose three novel metrics: ReXKG-NSC for assessing node similarity, ReXKG-AMS for evaluating edge distribution, and  ReXKG-SCS for measuring subgraph coverage across knowledge graphs.
These metrics allow for a global score comparison between models and against human radiologists, providing a comprehensive understanding of the model's performance.

Based on the knowledge graph and proposed metrics, we conduct a comprehensive analysis of both specialist and generalist report generation models, exploring the following questions and summarizing the main conclusions for each:

\noindent \textbf{Q1: Coverage of Entities.} How well do the generated reports cover essential entities such as anatomy and disorders?
\textcolor{black}{
Generalist models demonstrate broader coverage, capturing nearly 80\% of essential entities, yet they still fall short of matching the depth of radiologist-written reports, particularly in detailing medical devices.
}

\noindent \textbf{Q2: Coverage of Relationships Between Entities.} How comprehensively do the AI reports describe connections between different medical findings and their descriptions? \textcolor{black}{
All AI models show significant gaps compared to radiologist-written reports in capturing relationships between different entities, with MedVersa leading, achieving nearly 80\% coverage of the top 10\% subgraphs.
}

\noindent \textbf{Q3: Coverage of Concepts or Descriptors.} How detailed and comprehensive are the descriptions of disorders and anatomical features? 
\textcolor{black}{AI models tend to overfit specific concepts that appear frequently in the training data, resulting in less detailed and occasionally hallucinated descriptions.}

\noindent \textbf{Q4: Quantitative Measurements Coverage.} How frequently does the model provide quantified measurements of disorders? 
\textcolor{black}{
AI model's behavior in providing size descriptions correlates strongly with the frequency of size descriptions for specific disorders in the training data. 
}

\noindent \textbf{Q5: Specialist vs. Generalist Models.} What are the performance differences between specialist and generalist models?
\textcolor{black}{
Generalist models, trained on multiple modalities of data, demonstrate significantly enhanced radiology knowledge compared to specialist models. This suggests that exposure to a broader range of medical data and tasks contributes to a more comprehensive and accurate representation of radiological concepts and relationships.
}

\section{Knowledge Graph Construction}

\begin{figure*}[t]
    \centering
    \includegraphics[width=1\linewidth]{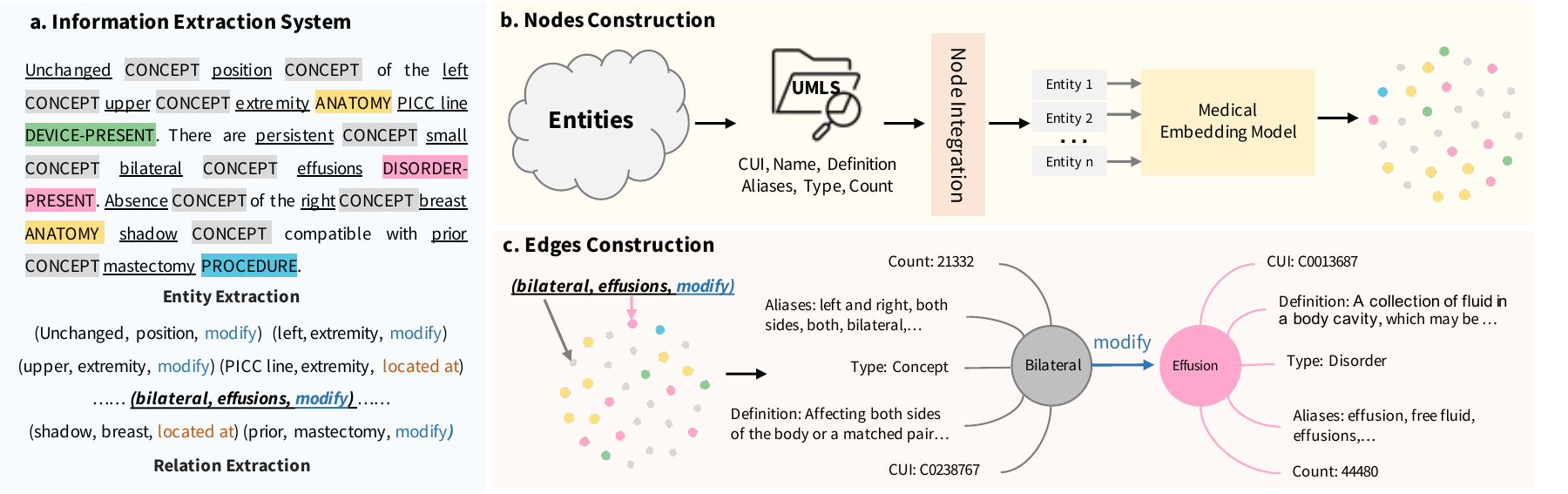}
    \vspace{-6pt}
    \caption{Overview of the proposed knowledge graph construction system \textbf{ReXKG}. (a) The information extraction system for entity and relation extraction. (b) The node construction pipeline. (c) Illustration of edge construction.}
    \label{fig:rexkg}
\end{figure*}

In this section, we present our system (\textbf{ReXKG}) for constructing a comprehensive knowledge graph from a large corpus of radiology reports, shown in Figure~\ref{fig:rexkg}.
We first define an information extraction schema tailored to the radiology domain, then once the entities and relationships are extracted, we proceed with the node construction pipeline to ensure data consistency and integrity. Finally, we integrate the information into the graph structure. 

\subsection{Information Extraction Schema}

\subsubsection{Definition.}
We define an entity as a continuous span of text that can include one or more adjacent words. Entities in our schema are categorized into six types as listed.  
\begin{itemize}
    \item \textbf{Anatomy}: anatomical structures within the body.
    \item \textbf{Disorder}: any abnormal findings or diseases identified within radiology reports.
    \item \textbf{Concept}: descriptors used to modify other entities, for example, "acute", "severe", and "increasing".
    \item \textbf{Device}: any instrument or apparatus used for medical purposes, for example, ``tube'', ``clip'', ``wire''.
    \item \textbf{Procedure}: medical procedures used to diagnose, measure, monitor, or treat conditions, such as ``sternotomy''.
    \item \textbf{Size}: measurements of disorders or anatomical structures, for example, ``3-mm''.
\end{itemize}
We define a relation as a directed edge between two entities. Following the previous work~\cite{jain2021radgraph}, our schema uses three relations as listed.
\begin{itemize}
    \item \textbf{Suggestive of}: source entity (e.g., findings) may suggest the presence of the target entity (e.g., a disease).
    \item \textbf{Located at}: source entity is located at the target entity.
    \item \textbf{Modify}: source entity modifies or provides additional information about the target entity.
\end{itemize}

\subsubsection{Entity and Relation Extraction.}
Given a set of radiology reports, we first annotate a subset using GPT-4~\cite{achiam2023gpt} to generate labeled entities and relations. 
The prompts used for annotation are provided in the appendix.
Based on the annotated data, we train the model using the Princeton University Relation Extraction system~(PURE) architecture~\cite{zhong2020frustratingly} to do Named Entity Recognition~(NER). This architecture employs a pipeline approach, decomposing the tasks of entity recognition and relation extraction into separate subtasks.
Once the model is trained, we apply it to the entire dataset to perform inference,  extracting all relevant entities and relations.

\subsection{Nodes Construction}
Following entity extraction, we employ a series of steps to remove noise, merge synonyms, and link entities to the Unified Medical Language System (UMLS)~\cite{bodenreider2004unified}.
First, we determine the entity type of each extracted entity based on the most frequently predicted type by the NER model to ensure consistency and accuracy. 
Next, we utilize ScispaCy~\cite{neumann2019scispacy} to retrieve UMLS attributes for each entity, such as Concept Unique Identifiers (CUI), Type Unique Identifiers (TUI), definitions, and aliases. 
Entities that cannot be mapped to a UMLS item are retained for further processing.
For entities identified as aliases of a specific term in UMLS, we normalize these entities by merging them into a single concept. 
For instance, entities such as ``pulmonary'' and  ``lung'' are normalized to their corresponding CUI C0024109.
Additionally, to ensure the compactness and unambiguity of nodes, for the multi-word entities, if all individual words of such an entity are predicted as separate nodes, the combined multi-word entity is not included as a node. The detailed algorithm is provided in the appendix.
Finally, we leverage medical language models to merge entities based on semantic similarity. 
Entities with an embedding similarity higher than a defined threshold are combined. 
This step enhances the graph's coherence by aggregating semantically similar concepts into single nodes.

\subsection{Edges Construction}
Initially, all relations are extracted from the dataset as triplets (source entity, target entity, relation). 
We merge different triplets with the same source and target entities based on node aliases. 
When two nodes are linked by multiple relation types, we retain the relation type most frequently predicted by the model. Finally, we filter the relations by ignoring triplets with a count less than $C$, a hyperparameter ensuring the reliability of the connections within the graph.

\section{Knowledge Graph Evaluation Metrics}

To evaluate knowledge graphs obtained from different models, we introduce three metrics that assess node similarity, edge distribution similarity, and subgraph coverage: \textbf{ReXKG-NSC} (Node Similarity Coefficient), \textbf{ReXKG-AMS} (Adjacency Matrix Similarity), and \textbf{ReXKG-SCS} (Subgraph Coverage Score).
In the following, we will first provide a preliminary definition of the knowledge graph and then detail the calculation methods for these metrics.

\subsection{Preliminary Definition}
Assume we have a knowledge graph with $N$ nodes and $M$ edges. The set of nodes is denoted as $V = \{ v_1, v_2, \ldots, v_N \}$. The weights of the nodes are represented as $W_V = \{ w_{v_1}, w_{v_2}, \ldots, w_{v_N} \}$, where  where $w_{v_i}$ corresponds to the frequency of node $v_i$ in the data.
The set of edges is denoted as $E = \{ e_1, e_2, \ldots, e_M \}$, where each edge $e_m$ connects a pair of nodes $(v_i, v_j)$.
The weights of the edges are represented as $W_E = \{ w_{e_1}, w_{e_2}, \ldots, w_{e_M} \}$, where $w_{e_m} = \text{count}(e_m)$.
Then, the adjacency matrix is defined as $A$, with $A_{ij} = w_{e_{ij}}$, representing the weight of the edge between nodes $v_i$ and $v_j$.

\subsection{KG Node Similarity Coefficient}
Let KG-GT represent the knowledge graph built from the ground truth reports, consisting of $N$ nodes. Similarly, let KG-Pred represent the knowledge graph built from the generated reports, consisting of $P$ nodes.
For each node $v_i$ in KG-GT, we identify the most similar node in KG-Pred, assigning a similarity score $s_i$ based on calculations from a medical language model.
The overall node similarity metric is then calculated as the average of these similarity scores across all nodes in KG-GT. This can be expressed as:
\begin{equation}
    \texttt{KG-NSC} = \frac{1}{N} \sum_{i=1}^{N} s_i.
\end{equation}

\subsection{KG Adjacency Matrix Similarity}
For each node $v_i$ in KG-GT, we identify the most similar node in KG-Pred. This allows us to map all edges in KG-Pred using the nodes from KG-GT, resulting in the creation of two adjacency matrices, $A_{Pred}$ and $A_{GT}$, both of the same size.
Where $A_{ij}$ represents the weight of the edge between nodes $i$ and $j$.
We use the Pearson correlation coefficient metrics to evaluate the coverage of relations in generated reports compared to the ground truth. 
The row weight $w_{r_i}$ is used as the weight, and the Pearson correlation coefficient as the value.
Here, for a given row $i$, the row weight is defined as 
$w_{r_i} = (\sum_j A_{ij}) / (\sum_i \sum_j A_{ij})$,
where $A_{ij}$ represents the element at row $i$, column $j$ of the adjacency matrix.
Thus, the adjacency matrix similarity can be expressed as:
\begin{equation}
\texttt{KG-AMS} = \frac{\sum_{i=1}^{N} \left( w_{r_i} \cdot \texttt{corr}(A_{Pred,i}, A_{GT,i}) \right)}{\sum_{i=1}^{N} w_{r_i}},
\end{equation}
where $\texttt{corr}(A_{Pred,i}, A_{GT,i})$ is the Pearson correlation coefficient between the $i$-th rows of $A_{Pred}$ and $A_{GT}$, and $w_{r_i}$ is the weight of all edges associated with the $i$-th row.

\subsection{KG Subgraph Coverage Score}
Let $\mathcal{S}$ be the set of all connected subgraphs in KG-GT up to a size of $k$ nodes. 
We quantify a model's ability to represent important subgraphs from KG-GT within KG-Pred, which can be expressed as:
\begin{equation}
\texttt{KG-SCS} = \frac{\sum_{i=1}^{K} I(S_i) \cdot P(S_i)}{\sum_{i=1}^{K} I(S_i)},
\end{equation}
where $K$ is the number of top important subgraphs considered. $I(S_i)$ is the importance score of each subgraph $S_i$ in KG-GT and $P(S_i)$ is the presence score in KG-Pred. Please refer to the appendix for detailed definitions.

\section{Experiments}
In this section, we present the dataset and models used in our analysis of AI-generated reports. 
Given the current limitations in model capabilities, with few models available for generating CT/MRI reports, our study primarily focuses on chest X-ray report analysis. 
However, the proposed ReXKG is versatile and applicable across various modalities and anatomical regions, as demonstrated in the appendix.

\subsection{Datasets}
\noindent \textbf{CheXpert Plus}: CheXpert Plus~\cite{chambon2024chexpert} is a dataset that pairs text and images, featuring 223,228 unique pairs of radiology reports and chest X-rays from 187,711 studies and 64,725 patients. Each patient may be linked to multiple studies, and each study may include several images.

\vspace{3pt}
\noindent \textbf{MIMIC CXR}: MIMIC-CXR~\cite{johnson2019mimic} is a large publicly available dataset of chest X-rays with free-text radiology reports. The dataset contains 377,110 images corresponding to 227,835 radiographic studies performed at the Beth Israel Deaconess Medical Center.  

\subsection{Experiments Settings}
To ensure a comprehensive analysis, we randomly split studies from CheXpert Plus into two parts: CheXpert Plus I (24,086 studies) and CheXpert Plus II (24,085 studies).
Additionally, we randomly select a subset from MIMIC-CXR with 24,085 studies for comparison.
We designate CheXpert Plus I as the benchmark for our study. 
This subset serves as the ground truth, upon which all model evaluations are conducted, inference tasks performed, and knowledge graphs constructed. 
Similarly, we can set CheXpert Plus II as the benchmark, with results provided in the appendix.
The knowledge graphs for comparison can be categorized into two groups based on the data source.

\vspace{3pt}
\noindent \textbf{Intra-Dataset Reports}: Intra-Dataset Reports are knowledge graphs built from real clinical datasets across different studies or centers. We use CheXpert Plus II and the selected MIMIC-CXR subset, which represent radiologist-written reports from various studies and centers, as benchmark baselines for comparison with AI-generated reports.

\vspace{3pt}
\noindent \textbf{Extra-Dataset Reports}: Extra-Dataset Reports are knowledge graphs constructed from AI-generated reports. To comprehensively evaluate AI performance, we assess various report generation models, including specialist models such as CvT2DistilGPT2~\cite{nicolson2023improving}, RGRG~\cite{tanida2023interactive}, and Swinv2-MIMIC~\cite{chambon2024chexpert}, as well as generalist models like CheXagent~\cite{chen2024chexagent}, RadFM~\cite{wu2023towards}, and MedVersa~\cite{zhou2024generalist}.
Here, specialist models are defined as those trained exclusively on chest X-ray report generation, whereas generalist models are large-scale models trained on various tasks.
Details of these models can be found in the appendix.

\subsection{Implementation Details}
For the Information Extraction Schema, we follow the approach described in \cite{jain2021radgraph}, utilizing the PURE framework \cite{zhong2020frustratingly}, which employs a pre-trained BERT model to obtain contextualized representations. These representations are then fed into a feedforward network to predict the probability distribution of entities, which subsequently serves as input for the relation model. The learning rate is set to 2e-5 during training.
We use MedCPT~\cite{jin2023medcpt} as the default medical language model for entity merging, with a merging threshold of 0.95. The threshold $C$ for edge construction is set to 5. 
The number of nodes in each subgraph is set to $k=2$, and the number of important subgraphs, $K$, is defined as 10\% of the total subgraphs in KG-GT.
For report generation inference, we use the code and checkpoints provided by the respective baseline models, focusing on the generation of the findings section.
All experiments are conducted on an NVIDIA A100 GPU.

\section{Results}
In this section, we present a comprehensive analysis of knowledge graphs generated from both intra-dataset reports (radiologist-written) and extra-dataset reports (AI-generated). 
Using CheXpert Plus I as our benchmark, we hypothesize that the knowledge graph generated from CheXpert Plus II will display similar nodes, edges, and distribution characteristics.
Such similarity would validate the consistency of our findings and underscore the reliability and quality of our proposed methods for constructing knowledge graphs.
Our analysis is structured around key questions that probe different aspects of report generation, from entity coverage to relationship comprehension, providing a multifaceted view of current AI models' capabilities.

\begin{table*}[tbh]
\label{tab:entity_coverage}
\small
\centering
\setlength{\tabcolsep}{2pt}
\begin{tabular}{ll|cccccc|cccc|c}
\toprule
\multirow{2}{*}{\textbf{Type}} & \multirow{2}{*}{\textbf{Models}}
&  \multicolumn{6}{|c}{\textbf{KG-NSC}} & \multicolumn{4}{|c}{\textbf{KG-AMS}} &   \multicolumn{1}{|c}{\textbf{KG-SCS}}\\
 & & Ana. & Dis. & Con. & Dev. & Pro. & All & Dis.Ana. & Dev.Ana. & Dis.Dis. & All  & k=2\\
\midrule
Intra-Dataset & CheXpert Plus II  & 0.974 & 0.967 & 0.970 & 0.958 & 0.977 & 0.970 & 0.966 & 0.981 & 0.988 & 0.971 & 0.981 \\
& MIMIC-CXR & 0.930 & 0.948 & 0.930 & 0.865 & 0.929 & 0.928 & 0.841 & 0.786 & 0.858 & 0.819 & 0.950 \\
\midrule
Specialist & CvT2DistilGPT2~\cite{nicolson2023improving} & 0.781 & 0.760 & 0.786 & 0.730 & 0.809  & 0.779 & 0.776 & 0.841 & 0.752 & 0.624 & 0.696 \\
& RGRG~\cite{tanida2023interactive} & 0.657 & 0.627 & 0.624 & 0.589 & 0.577 & 0.626 & 0.681 & 0.680 & 0.642 & 0.579 & 0.538 \\
& Swinv2-MIMIC~\cite{chambon2024chexpert} &  0.772 & 0.773 & 0.772 & 0.742 & 0.782 & 0.777  & 0.719 & 0.814 & 0.821 & 0.646 & 0.648 \\
\midrule
Generalist & CheXagent~\cite{chen2024chexagent} & 0.720 & 0.698 & 0.707 & 0.675 & 0.716 & 0.707 & 0.856 & \textbf{0.883} & 0.567 & 0.710 & 0.588\\
& RadFM~\cite{wu2023towards} & \textbf{0.817} & 0.829 & 0.796 & 0.732 & 0.777 & 0.800 & 0.725 & 0.695 & 0.538 & 0.601 & 0.733\\
& MedVersa~\cite{zhou2024generalist} & 0.807 & \textbf{0.830} & \textbf{0.801} & \textbf{0.754} & \textbf{0.818} & \textbf{0.804} & \textbf{0.859} & 0.843 & \textbf{0.894} & \textbf{0.748} & \textbf{0.806}\\
\bottomrule
\end{tabular}
\vspace{-3pt}
\caption{Knowledge graph comparison between CheXpert Plus I and Intra-Dataset or Extra-Dataset Reports. KG-NSC, KG-AMS, and KG-SCS scores are reported.
The best results are highlighted in boldface.
}
\end{table*}

\subsection{Q1: Coverage of Entities}

First, we explore the question: \textbf{How well do the AI-generated reports cover essential entities such as concepts, anatomy, disorders, devices, and procedures?}

\vspace{3pt} \noindent As shown in Table~\ref{tab:entity_coverage}, 
We compare the KG-NSC between CheXpert Plus I with other datasets and various report generation models. 
CheXpert Plus II and MIMIC-CXR, representing radiologist-written reports with similar and differing distributions of ground truth, exhibit high similarity across all entity types, with overall scores of 0.970 and 0.928. This high similarity demonstrates the reliability of the proposed metric and sets a high benchmark for AI models to match.
Among AI models, generalist models, particularly RadFM and MedVersa, exhibit broader coverage of essential entities compared to specialist models. This superior performance likely stems from their training on more diverse and large-scale datasets, enabling these models to generalize better and capture a wider range of medical entities.

When examining the results for each entity type, there is a noticeable gap in medical devices across all models.
This discrepancy may be attributed to the primary factor that models are exclusively trained on the MIMIC-CXR dataset, thus the models' predictions align more closely with MIMIC-CXR's distribution. 
However, there are inherent distribution differences between the CheXpert Plus and MIMIC-CXR datasets. CheXpert Plus includes some rare devices, such as the ``Impella'', which is mentioned only 15 times in the entire CheXpert Plus dataset. 
Additionally, varied terminology is used to describe the type of devices, such as ``keofeed'' for ``tubes''.

\subsection{Q2: Coverage of Relationships Between Entities}
Next, we investigate \textbf{How comprehensive is the coverage of relationships between entities?}

\noindent To evaluate the comprehensiveness of AI-generated reports in capturing relationships between entities, we employed the KG-AMS and KG-SCS metrics. Table~\ref{tab:entity_coverage} details the correlation between specific types of relationships: disorders with anatomy, devices with anatomy, and relationships between disorders.
MedVersa leads in the KG-AMS metric across most categories, particularly excelling in disorder-disorder and overall relationships. CheXagent, on the other hand, stands out in device-anatomy relationships, while RadFM shows balanced performance across various types of entity relationships. Despite these performances, there remains a significant gap compared to radiologist-written reports, highlighting areas for further improvement.
The KG-SCS metric (with $k$=2) offers additional insights into how well models capture important subgraphs or patterns within the knowledge graph. MedVersa covers 80.6\% of the important subgraphs, while RadFM covers over 73\%, indicating that while these models perform well, there is still room for enhancement in capturing complex relationships.

\begin{figure}[htb]
    \centering
    \includegraphics[width=1\linewidth]{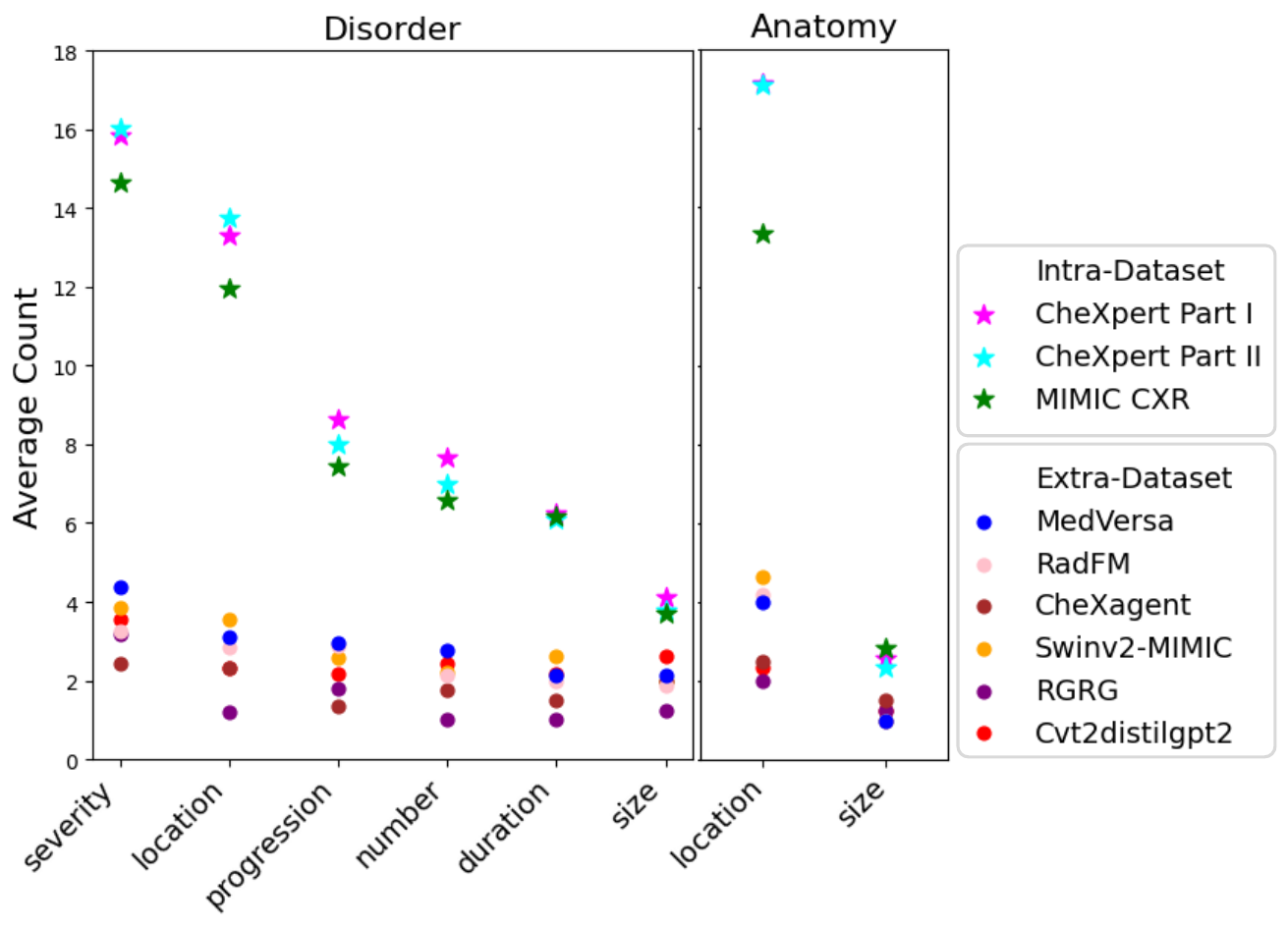}
    \vspace{-10pt}
    \caption{Average count of concept entities used to modify disorders and anatomy across different models.}
    \label{fig:concept_relation}
\end{figure}

\begin{figure*}[tbh]
    \centering
    \includegraphics[width=1\linewidth]{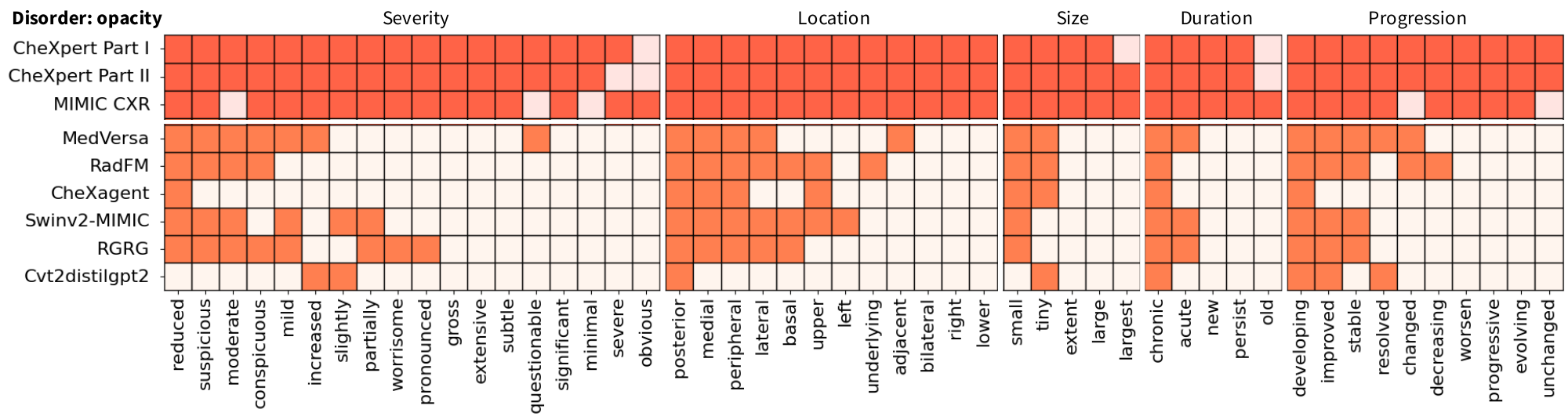}
    \vspace{-15pt}
    \caption{Detailed results of model predictions for given concepts related to specific disorders. Dark orange indicates the model predicts the relationship, while light orange indicates not. 
    }
    \label{fig:detailed_concept_relation}
\end{figure*}

\subsection{Q3: Comprehensiveness of Concepts} 
We further access the quality of content generated by AI models with the question: \textbf{How detailed and comprehensive are the descriptions of disorders and anatomical regions provided by the AI models?}

\vspace{3pt} \noindent This question is critical for applying AI models in clinical scenarios, where the ability to describe and differentiate the severity of diseases can directly impact diagnosis and treatment planning. 
To assess the depth and comprehensiveness with which disorders and anatomical regions are described, we utilize GPT-4 to classify all concept nodes within our knowledge graphs.  
These concepts are categorized into the following:
\begin{itemize}
    \item \textbf{Severity}: Describes how intense or severe the symptoms are, such as mild, moderate, or severe. 
    \item \textbf{Location}: Specifies where on or in the body the disorder manifests, such as left, right, bilateral, upper, lower, or specific organs or systems involved.
    \item \textbf{Duration}: Refers to how long the disorder or its symptoms have been present. (acute, chronic, transient)
    \item \textbf{Progression}: Indicates how the disorder changes over time, including progressive, stable, and regressive.
    \item \textbf{Size}: Relevant for physical abnormalities or tumors, indicating how large an affected area or lesion is. 
    \item \textbf{Number}: Describes how many lesions or abnormalities are present, such as single, multiple, or widespread.
\end{itemize}
Our analysis, depicted in Figure~\ref{fig:concept_relation}, shows that Intra-Dataset groups exhibit the highest similarity, with nearly identical counts for all category concepts used to modify disorders and anatomy.
 In contrast, AI models tend to underperform, especially in categories like ``severity'' and ``location''. Models often describe ``location'' for anatomy and ``severity'' for disorders, such as specifying ``left lung'' or ``mild edema'', but the range of terms they use for modification is limited.
Moreover, since all models perform inference without considering prior studies, concepts related to progression such as ``unchanged'' or ``improved'' may result from hallucinations
This issue arises partly because the training data often lack comprehensive, longitudinal information that accurately captures patient progression.
Additionally, some model training processes do not take into account the patient history or the continuity of patient data across multiple studies.

To gain a more detailed understanding, we selected several high-frequency disorders and the commonly used concepts to modify these disorders. 
One example is shown in Figure~\ref{fig:detailed_concept_relation}, the Intra-Dataset Reports's results exhibit complete coverage. In contrast, models tend to use concepts like ``moderate'' and ``mild'' but do not use terms ``severe'' or ``subtle'' for ``opacity''. 
We provide comprehensive detailed results in the appendix, from which we can observe that for some disorders, such as ``consolidation'', most models do not provide severity descriptions.
We also provide a barplot in the appendix showing the frequency of those concepts in MIMIC-CXR training set, an interesting observation is that the model's predictions are not linearly related to the frequency of appearance in the MIMIC-CXR training set. 
Instead, the model tends to overfit a specific synonym within a set of related concepts, and the selected concept varies for different disorders.

\begin{figure}[!t]
    \centering
    \includegraphics[width=1\linewidth]{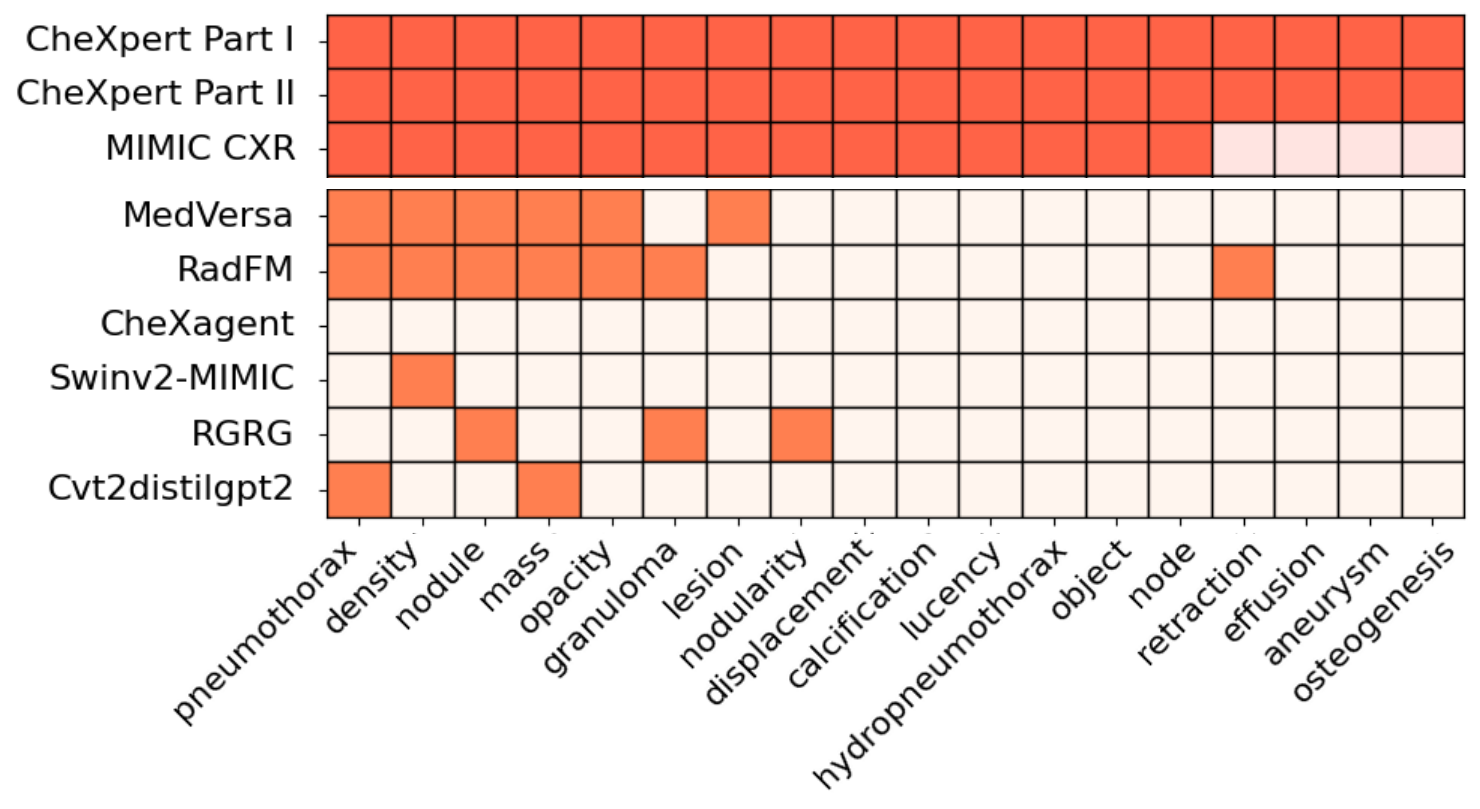}
    \vspace{-15pt}
    \caption{Detailed results of whether the model predicts specific size measurements for given disorders. Dark orange indicates the model predicts, while light indicates not.}
    \label{fig:detailed_size_relation}
\end{figure}

\subsection{Q4: Quantified Measurement} 
We then address the issue of quantification in the reports: \textbf{How frequently does the model provide quantified measurements of disorders and anatomical regions?}

\vspace{3pt} \noindent  This information is crucial for the deep analysis of images. For instance, disorders that consistently include size descriptions like ``3mm'' in the report might require the development of precise segmentation targets. On the other hand, some disorders that cannot be measured may only need bounding boxes during labeling. 
Based on the knowledge graph, AI research can easily identify which disorders can and should be segmented, thereby further promoting research on grounded report generation.

As shown in Figure~\ref{fig:detailed_size_relation}, we provide an overview of whether the models give detailed measurement descriptions for the target disorders. 
Both CheXpert Part I and CheXpert Part II consistently provide detailed descriptions for all target disorders, which highlights the real clinical requirements. 
However, most AI models show limited coverage, often failing to provide detailed descriptions for many conditions like calcification and effusion.
Relatively speaking, generalist models like RadFM and MedVersa cover a broader range of disorders. It is notable that CheXagent does not predict any size measurements for disorders but consistently provides size descriptions for devices such as tubes and lines.
We also provide the frequency of size descriptions for specific disorders in MIMIC-CXR training data in the appendix, as shown, the model's behavior in providing size descriptions correlates strongly with the frequency.

\begin{table}[!t]
\label{tab:metric}
\centering
\small
\setlength{\tabcolsep}{2pt}
\begin{tabular}{llcccccc}
\toprule
\textbf{Type} & \textbf{Models} & BLEU & BERT & Semb & RadG & RadC\\
\midrule
Specialist & CvT2DistilGPT2  & 0.123 & 0.262 & 0.286 & 0.119 & 1.585 \\
 & RGRG & \textbf{0.141} & \textbf{0.304} & 0.257 & \textbf{0.127} & 1.533\\
 & Swinv2-MIMIC  & 0.129 & 0.286 & 0.284 & 0.123 & 1.543  \\
\midrule
Generalist & CheXagent  & 0.102 & 0.299 & 0.294 & 0.124 & 1.510 \\
 & RadFM & 0.091 & 0.259 & 0.202 & 0.083 & 1.718 \\
 & MedVersa  &  0.116 & 0.300 & \textbf{0.315} & \textbf{0.127} & \textbf{1.483} \\
\bottomrule
\end{tabular}
\caption{Comparisons of both specialist and generalist models on CheXpert Plus. Metrics include BLEU, BERTScore (BERT), SembScore (Semb), RadGraph F1 (RadG), and RadCliQ-v1 (RadC).}
\end{table}

\subsection{Q5: Specialist vs. Generalist Models} 
Finally, we compare different types of AI models by asking: \textbf{What are the differences in performance between specialist models and generalist models?} 

\vspace{3pt} \noindent We summarize the score of different metrics on different models' predictions on the training set of CheXpert Plus finding sections.
Note that none of the models were trained using CheXpert Plus.
First, we observe that there is not a significant gap between the report-vs-report performance scores of specialist models and generalist models. This suggests that specialist models can perform well on specialist tasks. 
However, when comparing the models' knowledge coverage with that of radiologists, generalist models like RadFM and MedVersa show significantly broader node coverage. 
Note that here, all generalist models are trained on various tasks such as diagnosis, VQA, and report generation, but CheXagent only focuses on chest X-rays, while other generalist models include datasets from various modalities.
From this, we can conclude that including data from various modalities improves the models' prediction generalizability, especially in terms of entity coverage.
To develop medical AI systems that can interpret medical data and reason through complex problems at an expert radiologist level in real clinical scenarios, it is important to combine data from different modalities to broaden the models' knowledge base.

\vspace{-5pt}
\subsection{Ablation Studies}
We conduct ablation studies to examine the impact of different medical embedding models, similarity thresholds, and the number of reports on our proposed metrics. The results are presented in Table \ref{tab:ablation}.
First, we compare the performance of two medical embedding models, BioLoRD~\cite{remy2024biolord} and MedCPT~\cite{jin2023medcpt}, at different similarity thresholds. 
Our findings indicate that the choice of embedding model and threshold has a minimal effect on the extracted knowledge graph's quality. 
Both models perform robustly across different thresholds, with only slight variations in the KG-AMS metric.
We also investigate how the number of reports influences the quality of the resulting knowledge graph. As expected, the number of reports significantly affects the results. 
However, we observe that as the number of reports increases, the performance asymptotically approaches that of the full dataset. For instance, with 10,000 studies, we achieve a KG-NSC of 0.977 and a KG-AMS of 0.987, which closely matches the performance of the full dataset.

\begin{table}[!t]
\small
\centering
\setlength{\tabcolsep}{3pt}
\begin{tabular}{lllccc}
\toprule
Model & Threshold & \# Study  &  KG-NSC & KG-AMS & KG-SCS \\
\midrule
BioLoRD & 0.95  & 24,085 & \textbf{1.000} & \textbf{0.989} & \textbf{0.999}\\
BioLoRD & 0.90  & 24,085 & \textbf{1.000}  & 0.957 & 0.998 \\
MedCPT & 0.90  & 24,085 &  \textbf{1.000}  & 0.936 & 0.991 \\
\midrule
MedCPT & 0.95 & 100 & 0.769 & 0.858 & 0.585 \\
MedCPT & 0.95 & 1,000 & 0.923 & 0.933 & 0.864 \\
MedCPT & 0.95 & 10,000 & 0.977 & 0.987 & 0.997 \\
\bottomrule
\end{tabular}
\caption{Ablation studies on medical embedding models, similarity thresholds, and number of studies.}
\label{tab:ablation}
\end{table}

\section{Related Work}

Previous evaluations of radiology report generation models relied mainly on specific report-to-report metrics like FineRadScore~\cite{huang2024fineradscore}, RaTEScore~\cite{zhao2024ratescore}, RadFact~\cite{bannur2024maira}, CheXPrompt~\cite{juan2024towards}, and GREEN~\cite{ostmeier2024green}. These metrics, however, do not fully capture an in-depth understanding of the capabilities of current models.
Our work aims to address this limitation by leveraging knowledge graphs constructed from the report corpus. The standard pipeline for knowledge graph construction typically involves Named Entity Recognition~\cite{li2020survey}, Relation Extraction~\cite{pawar2017relation}, and Entity Resolution~\cite{christophides2020overview}. In the medical domain, the focus has primarily been on developing knowledge graphs based on complex medical systems such as electronic health records, medical literature, and clinical guidelines~\cite{rotmensch2017learning,finlayson2014building,bean2017knowledge}. However, in the specific context of radiology reports, most progress focuses on information extraction ~\cite{irvin2019chexpert,mcdermott2020chexpert++,peng2018negbio,smit2020chexbert,jain2021visualchexbert,jain2021radgraph,khanna2023radgraph2,delbrouck2024radgraph}, and have not yet led to the establishment of a comprehensive knowledge graph specifically tailored for radiology reports.
Few existing studies~\cite{kale2022knowledge, zhang2020radiology} related to knowledge graph construction heavily relied on manual annotation by radiologists, highlighting the need for more automated, scalable approaches in this field.

\section{Conclusion}
In this paper, we present ReXKG, a novel system for constructing comprehensive radiology knowledge graphs from medical reports, and introduce three metrics for evaluationg the similarity of nodes, distributions of edges, and coverage of subgraphs.
We conduct an in-depth analysis comparing AI-generated radiology reports to human-written reports.
Our research reveals that generalist models trained on various modalities offer broader coverage and enhanced radiology knowledge, yet they still fall short of the depth found in radiologist-written reports, particularly in the description and size measurements of disorders. 
Additionally, hallucinations related to prior studies are noticeable in model-generated reports, highlighting the need to incorporate longitudinal data in future model development.


\bibliography{aaai25}

\begin{thebibliography}{46}
\providecommand{\natexlab}[1]{#1}

\bibitem[{Achiam et~al.(2023)Achiam, Adler, Agarwal, Ahmad, Akkaya, Aleman,
  Almeida, Altenschmidt, Altman, Anadkat et~al.}]{achiam2023gpt}
Achiam, J.; Adler, S.; Agarwal, S.; Ahmad, L.; Akkaya, I.; Aleman, F.~L.;
  Almeida, D.; Altenschmidt, J.; Altman, S.; Anadkat, S.; et~al. 2023.
\newblock Gpt-4 technical report.
\newblock \emph{arXiv preprint arXiv:2303.08774}.

\bibitem[{Bannur et~al.(2024)Bannur, Bouzid, Castro, Schwaighofer, Bond-Taylor,
  Ilse, P{\'e}rez-Garc{\'\i}a, Salvatelli, Sharma, Meissen
  et~al.}]{bannur2024maira}
Bannur, S.; Bouzid, K.; Castro, D.~C.; Schwaighofer, A.; Bond-Taylor, S.; Ilse,
  M.; P{\'e}rez-Garc{\'\i}a, F.; Salvatelli, V.; Sharma, H.; Meissen, F.;
  et~al. 2024.
\newblock MAIRA-2: Grounded Radiology Report Generation.
\newblock \emph{arXiv preprint arXiv:2406.04449}.

\bibitem[{Bean et~al.(2017)Bean, Wu, Iqbal, Dzahini, Ibrahim, Broadbent,
  Stewart, and Dobson}]{bean2017knowledge}
Bean, D.~M.; Wu, H.; Iqbal, E.; Dzahini, O.; Ibrahim, Z.~M.; Broadbent, M.;
  Stewart, R.; and Dobson, R.~J. 2017.
\newblock Knowledge graph prediction of unknown adverse drug reactions and
  validation in electronic health records.
\newblock \emph{Scientific reports}, 7(1): 16416.

\bibitem[{Bodenreider(2004)}]{bodenreider2004unified}
Bodenreider, O. 2004.
\newblock The unified medical language system (UMLS): integrating biomedical
  terminology.
\newblock \emph{Nucleic acids research}, 32(suppl\_1): D267--D270.

\bibitem[{Chambon et~al.(2024)Chambon, Delbrouck, Sounack, Huang, Chen, Varma,
  Truong, Chuong, and Langlotz}]{chambon2024chexpert}
Chambon, P.; Delbrouck, J.-B.; Sounack, T.; Huang, S.-C.; Chen, Z.; Varma, M.;
  Truong, S.~Q.; Chuong, C.~T.; and Langlotz, C.~P. 2024.
\newblock CheXpert Plus: Hundreds of Thousands of Aligned Radiology Texts,
  Images and Patients.
\newblock \emph{arXiv preprint arXiv:2405.19538}.

\bibitem[{Chaves et~al.(2024)Chaves, Huang, Xu, Xu, Usuyama, Zhang, Wang, Xie,
  Khademi, Yang, Awadalla, Gong, Hu, Yang, Li, Gao, Gu, Wong, Wei, Naumann,
  Chen, Lungren, Yeung-Levy, Langlotz, Wang, and Poon}]{juan2024towards}
Chaves, J. M.~Z.; Huang, S.-C.; Xu, Y.; Xu, H.; Usuyama, N.; Zhang, S.; Wang,
  F.; Xie, Y.; Khademi, M.; Yang, Z.; Awadalla, H.~H.; Gong, J.; Hu, H.; Yang,
  J.; Li, C.; Gao, J.; Gu, Y.; Wong, C.; Wei, M.-H.; Naumann, T.; Chen, M.;
  Lungren, M.~P.; Yeung-Levy, S.; Langlotz, C.~P.; Wang, S.; and Poon, H. 2024.
\newblock Towards a clinically accessible radiology foundation model:
  open-access and lightweight, with automated evaluation.

\bibitem[{Chen et~al.(2024)Chen, Varma, Delbrouck, Paschali, Blankemeier,
  Van~Veen, Valanarasu, Youssef, Cohen, Reis et~al.}]{chen2024chexagent}
Chen, Z.; Varma, M.; Delbrouck, J.-B.; Paschali, M.; Blankemeier, L.; Van~Veen,
  D.; Valanarasu, J. M.~J.; Youssef, A.; Cohen, J.~P.; Reis, E.~P.; et~al.
  2024.
\newblock Chexagent: Towards a foundation model for chest x-ray interpretation.
\newblock \emph{arXiv preprint arXiv:2401.12208}.

\bibitem[{Christophides et~al.(2020)Christophides, Efthymiou, Palpanas,
  Papadakis, and Stefanidis}]{christophides2020overview}
Christophides, V.; Efthymiou, V.; Palpanas, T.; Papadakis, G.; and Stefanidis,
  K. 2020.
\newblock An overview of end-to-end entity resolution for big data.
\newblock \emph{ACM Computing Surveys (CSUR)}, 53(6): 1--42.

\bibitem[{Delbrouck et~al.(2024)Delbrouck, Chambon, Chen, Varma, Johnston,
  Blankemeier, Van~Veen, Bui, Truong, and Langlotz}]{delbrouck2024radgraph}
Delbrouck, J.-B.; Chambon, P.; Chen, Z.; Varma, M.; Johnston, A.; Blankemeier,
  L.; Van~Veen, D.; Bui, T.; Truong, S.; and Langlotz, C. 2024.
\newblock RadGraph-XL: A Large-Scale Expert-Annotated Dataset for Entity and
  Relation Extraction from Radiology Reports.
\newblock In \emph{Findings of the Association for Computational Linguistics
  ACL 2024}, 12902--12915.

\bibitem[{Finlayson, LePendu, and Shah(2014)}]{finlayson2014building}
Finlayson, S.~G.; LePendu, P.; and Shah, N.~H. 2014.
\newblock Building the graph of medicine from millions of clinical narratives.
\newblock \emph{Scientific data}, 1(1): 1--9.

\bibitem[{Huang et~al.(2024)Huang, Banerjee, Wu, Reis, and
  Rajpurkar}]{huang2024fineradscore}
Huang, A.; Banerjee, O.; Wu, K.; Reis, E.~P.; and Rajpurkar, P. 2024.
\newblock FineRadScore: A Radiology Report Line-by-Line Evaluation Technique
  Generating Corrections with Severity Scores.
\newblock \emph{arXiv preprint arXiv:2405.20613}.

\bibitem[{Irvin et~al.(2019)Irvin, Rajpurkar, Ko, Yu, Ciurea-Ilcus, Chute,
  Marklund, Haghgoo, Ball, Shpanskaya et~al.}]{irvin2019chexpert}
Irvin, J.; Rajpurkar, P.; Ko, M.; Yu, Y.; Ciurea-Ilcus, S.; Chute, C.;
  Marklund, H.; Haghgoo, B.; Ball, R.; Shpanskaya, K.; et~al. 2019.
\newblock Chexpert: A large chest radiograph dataset with uncertainty labels
  and expert comparison.
\newblock In \emph{Proceedings of the AAAI conference on artificial
  intelligence}, volume~33, 590--597.

\bibitem[{Jain et~al.(2021{\natexlab{a}})Jain, Agrawal, Saporta, Truong, Duong,
  Bui, Chambon, Zhang, Lungren, Ng, Langlotz, Rajpurkar, and
  Rajpurkar}]{jain2021radgraph}
Jain, S.; Agrawal, A.; Saporta, A.; Truong, S.; Duong, D. N. D.~N.; Bui, T.;
  Chambon, P.; Zhang, Y.; Lungren, M.; Ng, A.; Langlotz, C.; Rajpurkar, P.; and
  Rajpurkar, P. 2021{\natexlab{a}}.
\newblock Radgraph: Extracting clinical entities and relations from radiology
  reports.
\newblock In \emph{Proceedings of the Neural Information Processing Systems
  Track on Datasets and Benchmarks}, volume~1.

\bibitem[{Jain et~al.(2021{\natexlab{b}})Jain, Smit, Truong, Nguyen, Huynh,
  Jain, Young, Ng, Lungren, and Rajpurkar}]{jain2021visualchexbert}
Jain, S.; Smit, A.; Truong, S.~Q.; Nguyen, C.~D.; Huynh, M.-T.; Jain, M.;
  Young, V.~A.; Ng, A.~Y.; Lungren, M.~P.; and Rajpurkar, P.
  2021{\natexlab{b}}.
\newblock VisualCheXbert: addressing the discrepancy between radiology report
  labels and image labels.
\newblock In \emph{Proceedings of the Conference on Health, Inference, and
  Learning}, 105--115.

\bibitem[{Jin et~al.(2023)Jin, Kim, Chen, Comeau, Yeganova, Wilbur, and
  Lu}]{jin2023medcpt}
Jin, Q.; Kim, W.; Chen, Q.; Comeau, D.~C.; Yeganova, L.; Wilbur, W.~J.; and Lu,
  Z. 2023.
\newblock MedCPT: Contrastive Pre-trained Transformers with large-scale PubMed
  search logs for zero-shot biomedical information retrieval.
\newblock \emph{Bioinformatics}, 39(11): btad651.

\bibitem[{Johnson et~al.(2019)Johnson, Pollard, Berkowitz, Greenbaum, Lungren,
  Deng, Mark, and Horng}]{johnson2019mimic}
Johnson, A.~E.; Pollard, T.~J.; Berkowitz, S.~J.; Greenbaum, N.~R.; Lungren,
  M.~P.; Deng, C.-y.; Mark, R.~G.; and Horng, S. 2019.
\newblock MIMIC-CXR, a de-identified publicly available database of chest
  radiographs with free-text reports.
\newblock \emph{Scientific data}, 6(1): 317.

\bibitem[{Kale et~al.(2022)Kale, Bhattacharyya, Shetty, Gune, Shrivastava,
  Lawyer, and Biswas}]{kale2022knowledge}
Kale, K.; Bhattacharyya, P.; Shetty, A.; Gune, M.; Shrivastava, K.; Lawyer, R.;
  and Biswas, S. 2022.
\newblock Knowledge Graph Construction and Its Application in Automatic
  Radiology Report Generation from Radiologist's Dictation.
\newblock \emph{arXiv preprint arXiv:2206.06308}.

\bibitem[{Khanna et~al.(2023)Khanna, Dejl, Yoon, Truong, Duong, Saenz, and
  Rajpurkar}]{khanna2023radgraph2}
Khanna, S.; Dejl, A.; Yoon, K.; Truong, Q.~H.; Duong, H.; Saenz, A.; and
  Rajpurkar, P. 2023.
\newblock RadGraph2: Modeling Disease Progression in Radiology Reports via
  Hierarchical Information Extraction.
\newblock \emph{arXiv preprint arXiv:2308.05046}.

\bibitem[{Li et~al.(2020)Li, Sun, Han, and Li}]{li2020survey}
Li, J.; Sun, A.; Han, J.; and Li, C. 2020.
\newblock A survey on deep learning for named entity recognition.
\newblock \emph{IEEE Transactions on Knowledge and Data Engineering}, 34(1):
  50--70.

\bibitem[{Liu et~al.(2024)Liu, Tian, Chen, Song, and
  Zhang}]{liu2024bootstrapping}
Liu, C.; Tian, Y.; Chen, W.; Song, Y.; and Zhang, Y. 2024.
\newblock Bootstrapping Large Language Models for Radiology Report Generation.
\newblock In \emph{Proceedings of the AAAI Conference on Artificial
  Intelligence}, volume~38, 18635--18643.

\bibitem[{Liu, Tian, and Song(2023)}]{liu2023systematic}
Liu, C.; Tian, Y.; and Song, Y. 2023.
\newblock A systematic review of deep learning-based research on radiology
  report generation.
\newblock \emph{arXiv preprint arXiv:2311.14199}.

\bibitem[{McDermott et~al.(2020)McDermott, Hsu, Weng, Ghassemi, and
  Szolovits}]{mcdermott2020chexpert++}
McDermott, M.~B.; Hsu, T. M.~H.; Weng, W.-H.; Ghassemi, M.; and Szolovits, P.
  2020.
\newblock Chexpert++: Approximating the chexpert labeler for speed,
  differentiability, and probabilistic output.
\newblock In \emph{Machine Learning for Healthcare Conference}, 913--927. PMLR.

\bibitem[{Neumann et~al.(2019)Neumann, King, Beltagy, and
  Ammar}]{neumann2019scispacy}
Neumann, M.; King, D.; Beltagy, I.; and Ammar, W. 2019.
\newblock ScispaCy: fast and robust models for biomedical natural language
  processing.
\newblock \emph{arXiv preprint arXiv:1902.07669}.

\bibitem[{Nicolson et~al.(2023)}]{nicolson2023improving}
Nicolson, A.; et~al. 2023.
\newblock Improving chest X-ray report generation by leveraging warm starting.
\newblock \emph{Artificial intelligence in medicine}, 144: 102633.

\bibitem[{Ostmeier et~al.(2024)Ostmeier, Xu, Chen, Varma, Blankemeier,
  Bluethgen, Michalson, Moseley, Langlotz, Chaudhari
  et~al.}]{ostmeier2024green}
Ostmeier, S.; Xu, J.; Chen, Z.; Varma, M.; Blankemeier, L.; Bluethgen, C.;
  Michalson, A.~E.; Moseley, M.; Langlotz, C.; Chaudhari, A.~S.; et~al. 2024.
\newblock GREEN: Generative Radiology Report Evaluation and Error Notation.
\newblock \emph{arXiv preprint arXiv:2405.03595}.

\bibitem[{Papineni et~al.(2002)Papineni, Roukos, Ward, and
  Zhu}]{papineni2002bleu}
Papineni, K.; Roukos, S.; Ward, T.; and Zhu, W.-J. 2002.
\newblock Bleu: a method for automatic evaluation of machine translation.
\newblock In \emph{Proceedings of the 40th annual meeting of the Association
  for Computational Linguistics}, 311--318.

\bibitem[{Pawar, Palshikar, and Bhattacharyya(2017)}]{pawar2017relation}
Pawar, S.; Palshikar, G.~K.; and Bhattacharyya, P. 2017.
\newblock Relation extraction: A survey.
\newblock \emph{arXiv preprint arXiv:1712.05191}.

\bibitem[{Peng et~al.(2018)Peng, Wang, Lu, Bagheri, Summers, and
  Lu}]{peng2018negbio}
Peng, Y.; Wang, X.; Lu, L.; Bagheri, M.; Summers, R.; and Lu, Z. 2018.
\newblock NegBio: a high-performance tool for negation and uncertainty
  detection in radiology reports.
\newblock \emph{AMIA Summits on Translational Science Proceedings}, 2018: 188.

\bibitem[{Rajpurkar et~al.(2022)Rajpurkar, Chen, Banerjee, and
  Topol}]{rajpurkar2022ai}
Rajpurkar, P.; Chen, E.; Banerjee, O.; and Topol, E.~J. 2022.
\newblock AI in health and medicine.
\newblock \emph{Nature medicine}, 28(1): 31--38.

\bibitem[{Rajpurkar and Lungren(2023)}]{rajpurkar2023current}
Rajpurkar, P.; and Lungren, M.~P. 2023.
\newblock The current and future state of AI interpretation of medical images.
\newblock \emph{New England Journal of Medicine}, 388(21): 1981--1990.

\bibitem[{Reale-Nosei et~al.(2024)}]{reale2024vision}
Reale-Nosei, G.; et~al. 2024.
\newblock From vision to text: A comprehensive review of natural image
  captioning in medical diagnosis and radiology report generation.
\newblock \emph{Medical Image Analysis}, 103264.

\bibitem[{Remy, Demuynck, and Demeester(2024)}]{remy2024biolord}
Remy, F.; Demuynck, K.; and Demeester, T. 2024.
\newblock BioLORD-2023: semantic textual representations fusing large language
  models and clinical knowledge graph insights.
\newblock \emph{Journal of the American Medical Informatics Association},
  ocae029.

\bibitem[{Ridnik et~al.(2021)Ridnik, Ben-Baruch, Noy, and
  Zelnik-Manor}]{ridnik2021imagenet}
Ridnik, T.; Ben-Baruch, E.; Noy, A.; and Zelnik-Manor, L. 2021.
\newblock Imagenet-21k pretraining for the masses.
\newblock \emph{arXiv preprint arXiv:2104.10972}.

\bibitem[{Rotmensch et~al.(2017)Rotmensch, Halpern, Tlimat, Horng, and
  Sontag}]{rotmensch2017learning}
Rotmensch, M.; Halpern, Y.; Tlimat, A.; Horng, S.; and Sontag, D. 2017.
\newblock Learning a health knowledge graph from electronic medical records.
\newblock \emph{Scientific reports}, 7(1): 5994.

\bibitem[{Sanh et~al.(2019)Sanh, Debut, Chaumond, and
  Wolf}]{sanh2019distilbert}
Sanh, V.; Debut, L.; Chaumond, J.; and Wolf, T. 2019.
\newblock DistilBERT, a distilled version of BERT: smaller, faster, cheaper and
  lighter.
\newblock In \emph{NeurIPS EMC2 Workshop}.

\bibitem[{Smit et~al.(2020)Smit, Jain, Rajpurkar, Pareek, Ng, and
  Lungren}]{smit2020chexbert}
Smit, A.; Jain, S.; Rajpurkar, P.; Pareek, A.; Ng, A.~Y.; and Lungren, M.~P.
  2020.
\newblock CheXbert: combining automatic labelers and expert annotations for
  accurate radiology report labeling using BERT.
\newblock \emph{arXiv preprint arXiv:2004.09167}.

\bibitem[{Tanida et~al.(2023)Tanida, M{\"u}ller, Kaissis, and
  Rueckert}]{tanida2023interactive}
Tanida, T.; M{\"u}ller, P.; Kaissis, G.; and Rueckert, D. 2023.
\newblock Interactive and explainable region-guided radiology report
  generation.
\newblock In \emph{Proceedings of the IEEE/CVF Conference on Computer Vision
  and Pattern Recognition}, 7433--7442.

\bibitem[{Wu et~al.(2023)Wu, Zhang, Zhang, Wang, and Xie}]{wu2023towards}
Wu, C.; Zhang, X.; Zhang, Y.; Wang, Y.; and Xie, W. 2023.
\newblock Towards Generalist Foundation Model for Radiology by Leveraging
  Web-scale 2D\&3D Medical Data.
\newblock \emph{arXiv preprint arXiv:2308.02463}.

\bibitem[{Wu et~al.(2021{\natexlab{a}})Wu, Xiao, Codella, Liu, Dai, Yuan, and
  Zhang}]{wu2021cvt}
Wu, H.; Xiao, B.; Codella, N.; Liu, M.; Dai, X.; Yuan, L.; and Zhang, L.
  2021{\natexlab{a}}.
\newblock Cvt: Introducing convolutions to vision transformers.
\newblock In \emph{Proceedings of the IEEE/CVF international conference on
  computer vision}, 22--31.

\bibitem[{Wu et~al.(2021{\natexlab{b}})Wu, Agu, Lourentzou, Sharma, Paguio,
  Yao, Dee, Mitchell, Kashyap, Giovannini et~al.}]{wu2021chest}
Wu, J.~T.; Agu, N.~N.; Lourentzou, I.; Sharma, A.; Paguio, J.~A.; Yao, J.~S.;
  Dee, E.~C.; Mitchell, W.; Kashyap, S.; Giovannini, A.; et~al.
  2021{\natexlab{b}}.
\newblock Chest imagenome dataset for clinical reasoning.
\newblock \emph{arXiv preprint arXiv:2108.00316}.

\bibitem[{Yu et~al.(2023)Yu, Endo, Krishnan, Pan, Tsai, Reis, Fonseca, Lee,
  Abad, Ng et~al.}]{yu2023evaluating}
Yu, F.; Endo, M.; Krishnan, R.; Pan, I.; Tsai, A.; Reis, E.~P.; Fonseca, E. K.
  U.~N.; Lee, H. M.~H.; Abad, Z. S.~H.; Ng, A.~Y.; et~al. 2023.
\newblock Evaluating progress in automatic chest x-ray radiology report
  generation.
\newblock \emph{Patterns}, 4(9).

\bibitem[{Zhang et~al.(2019)Zhang, Kishore, Wu, Weinberger, and
  Artzi}]{zhang2019bertscore}
Zhang, T.; Kishore, V.; Wu, F.; Weinberger, K.~Q.; and Artzi, Y. 2019.
\newblock Bertscore: Evaluating text generation with bert.
\newblock \emph{arXiv preprint arXiv:1904.09675}.

\bibitem[{Zhang et~al.(2020)Zhang, Wang, Xu, Yu, Yuille, and
  Xu}]{zhang2020radiology}
Zhang, Y.; Wang, X.; Xu, Z.; Yu, Q.; Yuille, A.; and Xu, D. 2020.
\newblock When radiology report generation meets knowledge graph.
\newblock In \emph{Proceedings of the AAAI conference on artificial
  intelligence}, volume~34, 12910--12917.

\bibitem[{Zhao et~al.(2024)Zhao, Wu, Zhang, Zhang, Wang, and
  Xie}]{zhao2024ratescore}
Zhao, W.; Wu, C.; Zhang, X.; Zhang, Y.; Wang, Y.; and Xie, W. 2024.
\newblock RaTEScore: A Metric for Radiology Report Generation.
\newblock \emph{medRxiv}, 2024--06.

\bibitem[{Zhong and Chen(2021)}]{zhong2020frustratingly}
Zhong, Z.; and Chen, D. 2021.
\newblock A frustratingly easy approach for entity and relation extraction.
\newblock In \emph{Proceedings of the 2021 Conference of the North American
  Chapter of the Association for Computational Linguistics: Human Language
  Technologies}, 50--61.

\bibitem[{Zhou et~al.(2024)Zhou, Adithan, Acosta, Topol, and
  Rajpurkar}]{zhou2024generalist}
Zhou, H.-Y.; Adithan, S.; Acosta, J.~N.; Topol, E.~J.; and Rajpurkar, P. 2024.
\newblock A Generalist Learner for Multifaceted Medical Image Interpretation.
\newblock \emph{arXiv preprint arXiv:2405.07988}.

\end{thebibliography}

\clearpage 

\appendix

\section{Appendix}
\renewcommand{\thetable}{A\arabic{table}}
\renewcommand{\thefigure}{A\arabic{figure}}
\setcounter{table}{0}
\setcounter{figure}{0}

\subsection{Prompt for Entity Extraction}
\begin{grayquote}
    You are a radiologist performing clinical term extraction from the FINDINGS and IMPRESSION sections in the radiology report. Here a clinical term can be in [\texttt{anatomy}, \texttt{disorder\_present}, \texttt{disorder\_notpresent}, \texttt{procedure}, \texttt{device\_present}, \texttt{device\_notpresent}, \texttt{size}, \texttt{concept}]. \texttt{anatomy} refers to the anatomical body. \texttt{disorder\_present} refers to findings or diseases that are present according to the sentence. \texttt{disorder\_notpresent} refers to findings or diseases that are not present according to the sentence. \texttt{procedure} refers to procedures used to diagnose, measure, monitor, or treat problems. \texttt{device\_present} refers to any instrument, apparatus for medical purpose that are present according to the sentence. \texttt{device\_notpresent} refers to any instrument, apparatus for medical purpose that are not present according to the sentence. \texttt{size} refers to the measurement of disorders or anatomy, for example, \texttt{3mm}, \texttt{4x5 cm}. \texttt{concept} refers to descriptors such as \texttt{acute} or \texttt{chronic}, \texttt{large}, size or severity, or other modifiers, or descriptors of anatomy being normal. For example, right pleural effusion, \texttt{right} should be a \texttt{concept}, and \texttt{pleural} should be  \texttt{anatomy} and \texttt{effusion} should be \texttt{disorder-present} or \texttt{disorder-notpresent}. For example, normal cardiomediastinal silhouette. \texttt{normal} and \texttt{silhouette} should be \texttt{concept}, \texttt{cardiomediastinal} should be \texttt{anatomy}. Please extract terms one word at a time whenever possible, avoiding phrases. Note that terms like \texttt{no} and \texttt{no evidence of} are not considered entities. Given a list of radiology sentence input in the format: \textless{}Input\textgreater{}\textless{}sentence\textgreater{}\textless{}sentence\textgreater{}\textless{}/Input\textgreater{} Please reply with the JSON format following template: \{\textless{}sentence\textgreater{}\:\{\texttt{entity}:\texttt{entity type}, \texttt{entity}:\texttt{entity type}\}, \textless{}sentence\textgreater{}\:\{\texttt{entity}:\texttt{entity type}, \texttt{entity}:\texttt{entity type}\}\}.
\end{grayquote}

\subsection{Prompt for Relation Extraction}
\begin{grayquote}
    You are a radiologist performing relation extraction of entities from the FINDINGS and IMPRESSION sections in the radiology report. Here a clinical term can be in [\texttt{anatomy}, \texttt{disorder\_present}, \texttt{disorder\_notpresent}, \texttt{procedures}, \texttt{procedures}, \texttt{concept}, \texttt{devices\_present}, \texttt{devices\_notpresent}]. And the relation can be in [\texttt{modify}, \texttt{located\_at}, \texttt{suggestive\_of}].  \texttt{suggestive\_of} means the source entity (findings) may suggest the target entity (disease). \texttt{located\_at} means the source entity is located at the target entity. \texttt{modify} denotes the source entity modifies the target entity. Every time there is a \texttt{modify} relationship between concept and anatomy, the direction should be concept $\rightarrow$ anatomy. For example, paranasal sinuses are clear: source entity \texttt{clear} (concept), modify target entity \texttt{paranasal sinuses} (anatomy). For example, acute hemorrhage: source entity \texttt{acute} (concept), modify target entity \texttt{hemorrhage}. Given a piece of radiology text input in the JSON format: \{\texttt{sentence}:\{\texttt{entity}:\texttt{entity\_type}\}, \texttt{sentence}:\{\texttt{entity}:\texttt{entity\_type}\}\}. 
    Please reply with the following JSON format: \{\texttt{sentence}:[\{source entity:\texttt{target entity}, relation:\texttt{relation}\}, \{source entity:\texttt{target entity}, relation:\texttt{relation}\}\}
\end{grayquote}

\subsection{Algorithm for Node Construction}

\begin{algorithm}[htb]
\caption{Node Integration}
\label{alg:node_construction}
\begin{algorithmic}[1]
\REQUIRE $E$: list of entities
\REQUIRE $C$: count threshold
\REQUIRE $n$: maximum number of words in an entity

\STATE Initialize $A \leftarrow \emptyset$ \COMMENT{Set of initial nodes}

\STATE Group $E$ by word count and filter by $C$

\FOR{each $k$ from 1 to $n$}
    \FOR{each $e\in E$ with $k$ words}
        \IF{$k == 1$}
            \STATE Add $e$ to set $A$
        \ELSE
            \IF{$e$ can merge from nodes in $A$}
                \STATE Pass
            \ELSE
                \STATE Add $e$ to set $A$
            \ENDIF
        \ENDIF
    \ENDFOR
\ENDFOR

\RETURN $A$ \COMMENT{Set of nodes}
\end{algorithmic}
\end{algorithm}

\begin{table*}[tbh]
\small
\centering
\setlength{\tabcolsep}{4pt}
\begin{tabular}{lll|cccccc|cccc|c}
\toprule
\multirow{2}{*}{\textbf{Dataset}} & \multirow{2}{*}{\textbf{Source}} & \multirow{2}{*}{\textbf{Target}}
&  \multicolumn{6}{|c}{\textbf{KG-NSC}} & \multicolumn{4}{|c}{\textbf{KG-AMS}} & \multicolumn{1}{|c}{\textbf{KG-SCS}} \\
 & &  & Ana. & Dis. & Con. & Dev. & Pro. & All & Dis.Ana. & Dev.Ana. & Dis.Dis. & All & k=2 \\
\midrule
CT-RATE & Part I & Part II  & 0.977 & 0.971 & 0.984 & 0.955 & 0.973 & 0.978 & 0.997 & 0.914 & 0.972 & 0.974 & 0.999 \\
CT-RATE & Part II & Part I  & 0.982 & 0.968 & 0.977 & 0.977 & 0.991 & 0.977 & 0.997 & 0.974 & 0.993 & 0.948 & 0.998 \\
\midrule
MIMIC-IV Head CT & Part I & Part II  & 0.986 & 0.976 & 0.986 & 0.976 & 0.987 & 0.984 &0.989 & 0.986 & 0.994 & 0.993 & 0.999 \\
MIMIC-IV Head CT & Part II & Part I  & 0.981 & 0.977 & 0.983 & 0.952 & 0.972 & 0.980 &0.994 & 0.987 & 0.996& 0.987 & 0.999 \\
\bottomrule
\end{tabular}
\caption{Knowledge graph comparison on CT-RATE and MIMIC-IC Head CT datasets. KG-NSC, KG-AMS, and KG-SCS scores are reported.
The best results are highlighted in boldface.
}
\label{tab:entity_coverage}
\end{table*}

\begin{table*}[tbh]
\small
\centering
\setlength{\tabcolsep}{2pt}
\begin{tabular}{ll|cccccc|cccc|c}
\toprule
\multirow{2}{*}{\textbf{Type}} & \multirow{2}{*}{\textbf{Models}}
&  \multicolumn{6}{|c}{\textbf{KG-NSC}} & \multicolumn{4}{|c}{\textbf{KG-AMS}} & \multicolumn{1}{|c}{\textbf{KG-SCS}}\\
 & & Ana. & Dis. & Con. & Dev. & Pro. & All & Dis.Ana. & Dev.Ana. & Dis.Dis. & All & k=2 \\
\midrule
Intra-Dataset & CheXpert Plus I  & 0.970 & 0.967 & 0.974 & 0.980 & 0.980 & 0.973 & 0.954 & 0.983 & 0.985 & 0.968 & 0.997\\
& MIMIC-CXR &  0.920 & 0.956 & 0.936 & 0.882 & 0.938 & 0.932 & 0.844 & 0.807 & 0.849 & 0.832 & 0.952 \\
\midrule
Specialist & CvT2DistilGPT2~\cite{nicolson2023improving} &  0.776 & 0.772 & 0.787 & 0.747 & 0.806 & 0.781 & 0.751 & 0.846 & 0.692 & 0.644 & 0.664 \\
& RGRG~\cite{tanida2023interactive} &  0.664 & 0.636 & 0.618 & 0.597 & 0.568 & 0.626 & 0.612 & 0.681 & 0.725 & 0.578 & 0.529\\
& Swinv2-MIMIC~\cite{chambon2024chexpert}& 0.790 & 0.792 & 0.774 & 0.732 & 0.812 & 0.780 & 0.690 & 0.811 & 0.719 & 0.660 & 0.625\\
\midrule
Generalist & CheXagent~\cite{chen2024chexagent} &  0.715 & 0.696 & 0.698 & 0.686 & 0.718 & 0.702 & 0.779 & \textbf{0.877} & 0.566 & 0.711 & 0.555\\
& RadFM~\cite{wu2023towards} & \textbf{0.804 }& \textbf{0.831} & 0.788 & 0.728 & 0.765 & 0.792 & 0.681 & 0.704 & 0.613 & 0.615 & 0.635\\
& MedVersa~\cite{zhou2024generalist} &  \textbf{0.804} & 0.824 & \textbf{0.800} & \textbf{0.750} & \textbf{0.813} & \textbf{0.802} & \textbf{0.800 }& 0.851 &\textbf{ 0.893 }& \textbf{0.723} & \textbf{0.709}\\
\bottomrule
\end{tabular}
\caption{Knowledge graph comparison between CheXpert Plus II and Intra-Dataset or Extra-Dataset Reports. KG-NSC, KG-AMS, and KG-SCS scores are reported.
The best results are highlighted in boldface.
}
\label{tab:entity_coverage_chexpertplus2}
\end{table*}

\subsection{KG Subgraph Coverage Score}
Let $\mathcal{S} = \{S_1, S_2, ..., S_L\}$ be the set of all connected subgraphs in KG-GT with the size of $k$ nodes. For each subgraph $S_i$, we compute an importance score $I(S_i)$ based on the frequency of occurrence and total edge weights:
\begin{equation}
I(S_i) = \sum_{v \in V(S_i)} w_v \cdot \sum_{e \in E(S_i)} w_e,
\end{equation}
where $V(S_i)$ and $E(S_i)$ denote the vertex and edge sets of $S_i$ respectively, and $w_v$ and $w_e$ are the corresponding node and edge weights.
For each subgraph $S_i$ in KG-GT, we compute a presence score $P(S_i)$ in KG-Pred:
\begin{equation}
P(S_i) = \frac{1}{2} \left(\frac{|E(S'_i)|}{|E(S_i)|} + \frac{\sum_{v \in V(S_i)} s_v}{|V(S_i)|}\right),
\end{equation}
where $S'_i$ is the corresponding subgraph in KG-Pred, $|E(.)|$ and $|V(.)|$ denote the number of edges and vertices respectively, and $s_v$ is the similarity score between matched nodes as defined in the KG-NSC section.
The Subgraph Coverage Score is then calculated as:
\begin{equation}
\texttt{KG-SCS} = \frac{\sum_{i=1}^{K} I(S_i) \cdot P(S_i)}{\sum_{i=1}^{K} I(S_i)},
\end{equation}
where $K$ is the number of top important subgraphs considered,  $I(S_i)$ is the normalized importance score of subgraph $S_i$ among the selected $K$ subgraphs.

\subsection{Report Genertaion Models}

\begin{itemize}
    \item \textbf{CvT2DistilGPT2}~\cite{nicolson2023improving}: The model adopts the Convolutional Vision Transformer (CvT)~\cite{wu2021cvt} pre-trained on ImageNet-21K~\cite{ridnik2021imagenet} for the visual encoder and the Distilled Generative Pre-trained Transformer 2 (DistilGPT2)~\cite{sanh2019distilbert} for the text decoder. We use the released checkpoint trained on the MIMIC-CXR dataset.
    \item \textbf{RGRG}\cite{tanida2023interactive}: The model employs an anatomy-based object detector, fine-tuned on the Chest ImaGenome dataset~\cite{wu2021chest}, which identifies 29 annotated anatomical regions. These regional visual features are then used to guide the generation of detailed and clinically relevant radiology reports
    \item \textbf{Swinv2-MIMIC}\cite{chambon2024chexpert}: The model is proposed as a baseline model for report generation on the CheXpert Plus dataset~\cite{chambon2024chexpert}. It builds upon the Swin Transformer architecture, and for our experiments, we use the released checkpoint trained on the MIMIC-CXR findings dataset.
    \item \textbf{CheXagent}\cite{chen2024chexagent}: The model is trained on the CheXinstruct dataset, which utilizes a clinical large language model for parsing radiology reports, a vision encoder for CXR representation, and a network that bridges vision and language modalities. 
    \item \textbf{RadFM}\cite{wu2023towards}: The model is a radiology foundation model trained on large-scale multi-modal medical datasets, which enables the integration of text input interleaved with 2D or 3D medical scans to generate responses for diverse radiologic tasks.
    \item \textbf{MedVersa}\cite{zhou2024generalist}: The model is a versatile model trained on large-scale medical data across multiple modalities and tasks, which supports multimodal inputs, outputs, and on-the-fly task specification.
\end{itemize}

\begin{figure*}[!t]
    \centering
    \includegraphics[width=1\linewidth]{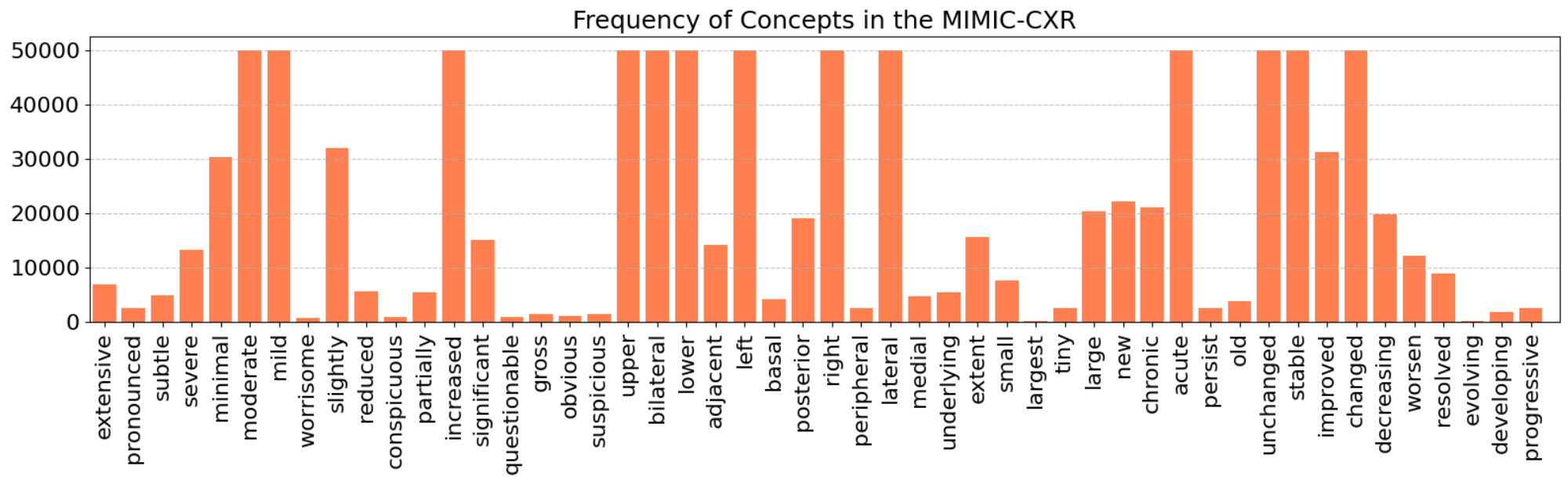}
    \caption{Frequency of concepts used to modify different disorders in the training set MIMIC-CXR.}
    \label{fig:mimiccxr_concept_frequency}
\end{figure*}

\subsection{Evaluation Metrics}
\begin{itemize}
    \item \textbf{BLEU}~\cite{papineni2002bleu} evaluates the precision of generated text by comparing n-gram overlap between the generated report and reference reports.
    \item \textbf{BERTScore}~\cite{zhang2019bertscore} employs a pre-trained BERT model to compute the similarity of word embeddings between candidate and reference texts.
    \item \textbf{SembScore}~\cite{smit2020chexbert} refers to the CheXbert labeler vector similarity. This method uses a 14-dimensional vector to indicate the presence of 13 common symptoms and the ``no finding" observation for each report, then calculates the cosine similarity between these vectors.
    \item \textbf{RadGraph F1}~\cite{jain2021radgraph} extracts radiology entities and relations specifically for Chest X-ray modality and computes the F1 score at the entity level.
    \item \textbf{RadCliQ-v1}~\cite{yu2023evaluating} is a composite metric that incorporates BLEU, BERTScore, SembScore, and RadGraph F1.
\end{itemize}

\subsection{Demonstration of ReXKG on various modalities}
The proposed knowledge graph construction system is versatile and can be applied across various modalities and anatomical regions. 
We further demonstrate its effectiveness on CT-RATE and MIMIC-IV Head CT reports, similar to the chest x-ray experiments. 
For these studies, we randomly split the target dataset into two equal parts and compared the knowledge graphs constructed from each subset.
\begin{itemize}
    \item \textbf{CT-RATE}: CT-RATE consists of 25,692 non-contrast chest CT volumes, expanded to 50,188 through various reconstructions, from 21,304 unique patients, along with corresponding radiology text reports. Here, we split the studies into two parts, Part I and Part II.
    \item \textbf{MIMIC-IV Head CT}: MIMIC-IV notes include reports from various modalities. Here we select the reports from head CT, including 101,633 studies, and split them into two parts, Part I and Part II.
\end{itemize}
The results in Table~\ref{tab:entity_coverage} show that when two corpora used for knowledge graph construction are of similar quality, the scores are consistently high. This indicates that the metrics are robust and suitable for evaluating knowledge graphs across various modalities.

\subsection{Results with CheXpert Plus II as benchmark}
Here, we set CheXpert Plus II as the benchmark and reproduce all the experiments, with results provided in Table~\ref{tab:entity_coverage_chexpertplus2}. As shown, the experimental results are consistent with those presented in the results section using CheXpert Plus I as the benchmark.

\subsection{Analysis of the concept used to modify disorders }
\begin{figure}[!t]
    \centering
    \includegraphics[width=1\linewidth]{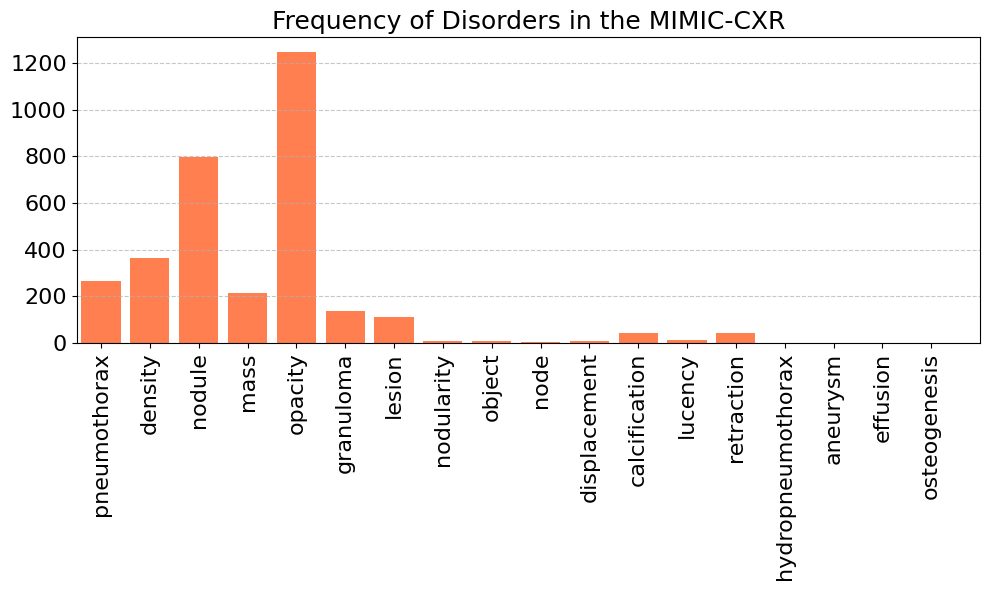}
    \caption{Frequency of size measurement for different disorders in the training set MIMIC-CXR.}
    \label{fig:mimiccxr_concept_size_frequency}
\end{figure}

\begin{figure*}[htb]
    \centering
    \includegraphics[width=1\linewidth]{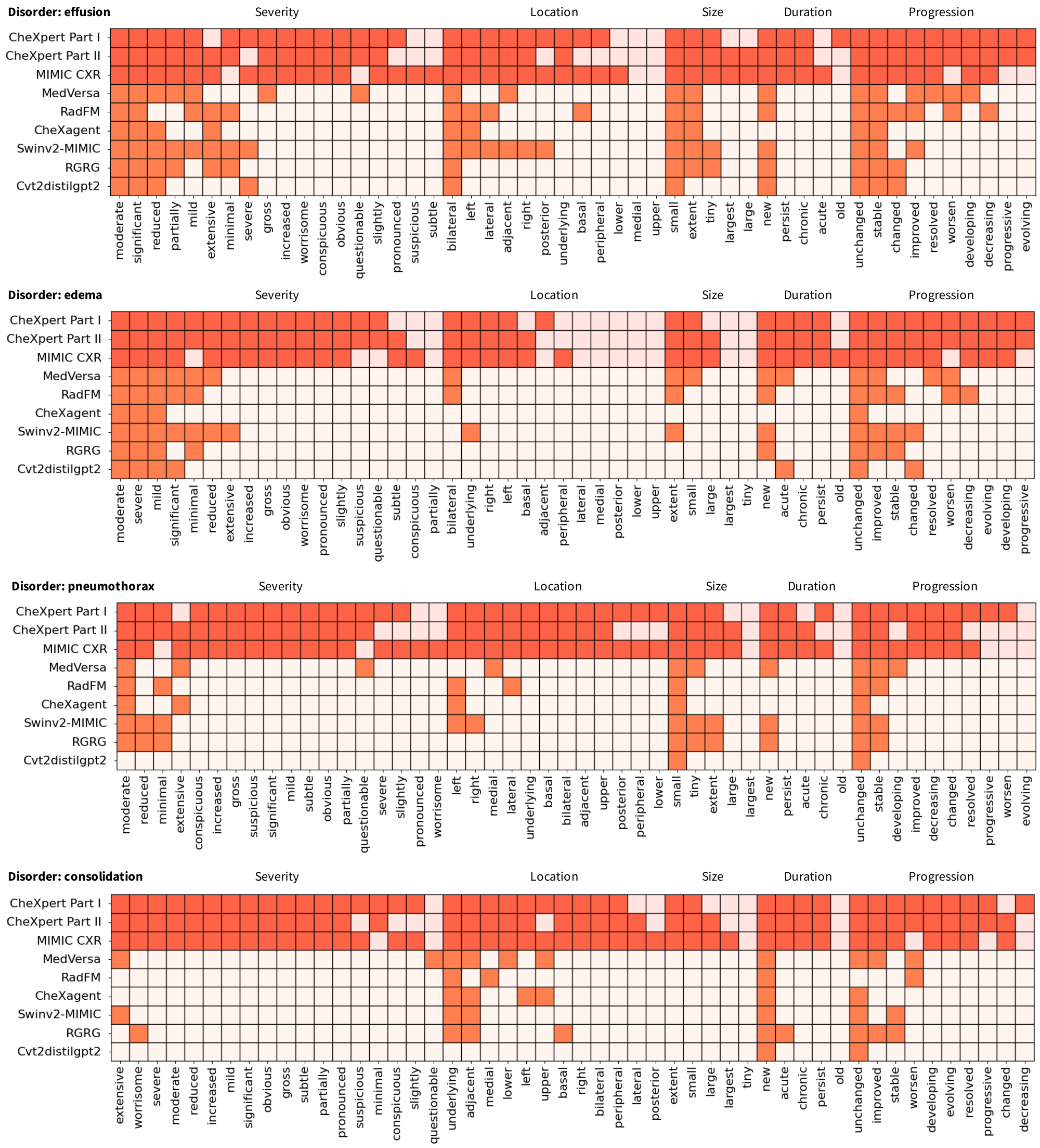}
    \caption{Detailed results of model predictions.}
    \label{fig:mimiccxr_concept_appendex1}
\end{figure*}

\begin{figure*}[htb]
    \centering
    \includegraphics[width=1\linewidth]{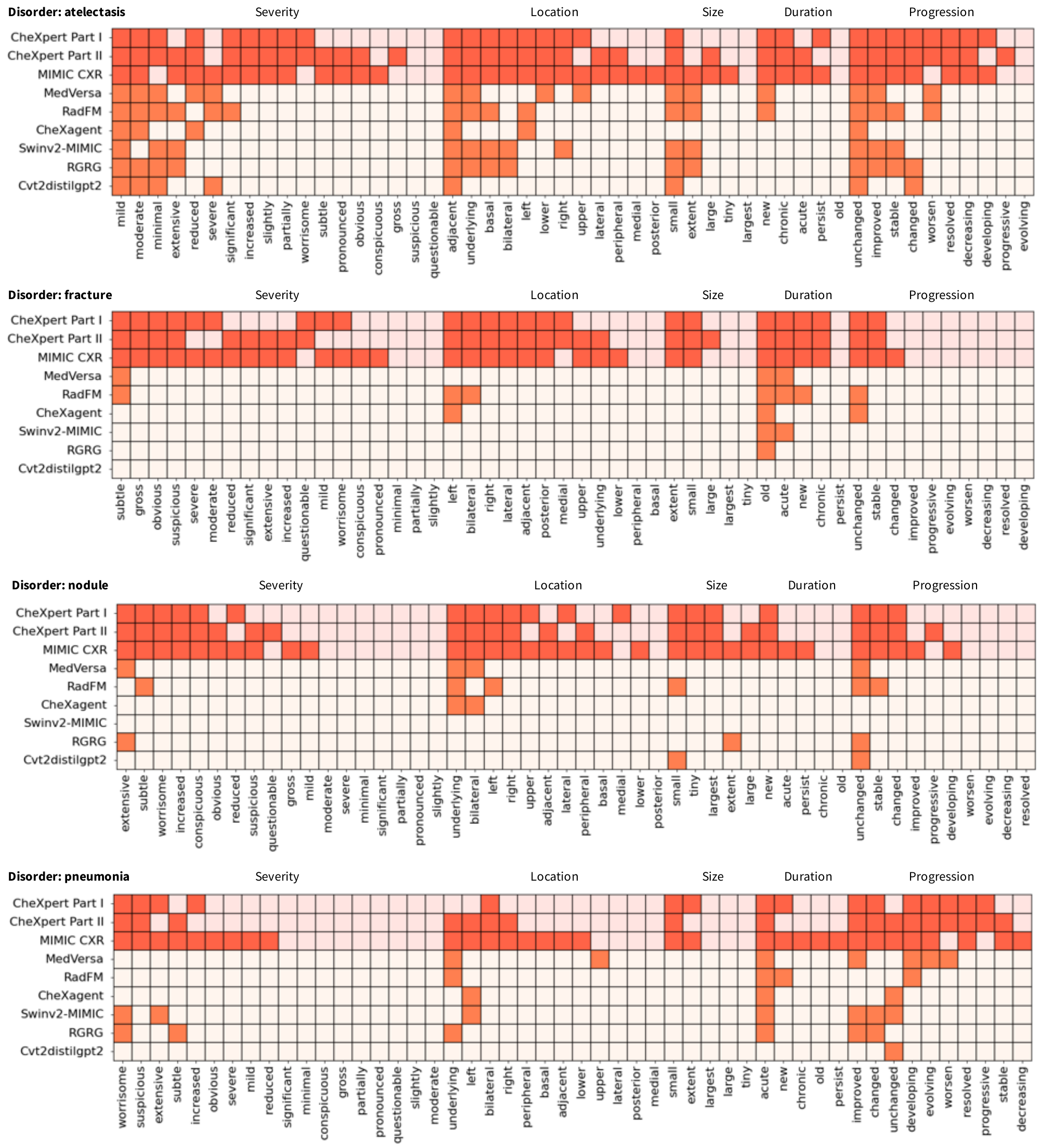}
    \caption{Detailed results of model predictions.}
    \label{fig:mimiccxr_concept_appendex2}
\end{figure*}

Figure~\ref{fig:mimiccxr_concept_frequency} illustrates the frequency distribution of the analyzed concepts in the MIMIC-CXR training set.
Figure~\ref{fig:mimiccxr_concept_size_frequency} depicts the frequency of size descriptions for specific disorders in the MIMIC-CXR training data.
Figure~\ref{fig:mimiccxr_concept_appendex1} and Figure~\ref{fig:mimiccxr_concept_appendex2} provide comprehensive results on high-frequency disorders and the commonly used concepts to modify these disorders across different models.

\end{document}